\documentclass{article}
\usepackage{graphicx}
\usepackage{amssymb}
\usepackage{epstopdf}
\usepackage{color}
\usepackage[ruled,linesnumbered]{algorithm2e}
\usepackage{pdflscape}
\usepackage{url}
\usepackage{xfrac}

\usepackage{natbib}
\usepackage{amsmath}
\usepackage[table]{xcolor}
\usepackage{setspace}
\usepackage[margin=1in]{geometry}
\usepackage{ulem}

\begin{document}
\onehalfspacing

\title{Improving Naive Bayes for Regression 
with Optimised Artificial Surrogate Data
}
\author{Michael Mayo and Eibe Frank \\
        Department of Computer Science\\
        University of Waikato\\
        Hamilton, New Zealand
}

\maketitle

\begin{abstract}
Can we evolve better training data for machine learning algorithms?
To investigate this question
we use population-based optimisation algorithms to 
generate artificial surrogate training data for naive Bayes for regression.
We demonstrate that the generalisation performance
of naive Bayes for regression models
is enhanced
by training them on the artificial data
as opposed to the real data.
These results are important for two reasons. 
Firstly, naive Bayes models are simple and interpretable
but frequently underperform compared to more complex ``black box''
models, and therefore new methods of enhancing accuracy are called for.
Secondly, the idea of using the real training data
indirectly in the construction of the artificial training data,
as opposed to directly for model training, 
is a novel twist on the usual machine learning paradigm.
\end{abstract}

\subsubsection*{Keywords}
particle swarm optimisation,
naive Bayes,
regression,
artificial data

\section{Introduction}

The typical pipeline for a supervised machine learning 
project involves firstly the collection of
a significant sample of labelled examples 
typically referred to as training data.
Depending on whether the labels are continuous or categorical, 
the supervised learning task is known as regression or classification 
respectively.
Next, once the training data is sufficiently clean and complete,
it is used to directly build a predictive model using
the machine learning algorithm of choice.
The predictive model is then used to label new unlabelled examples,
and if the labels of the new examples are known a priori by the user 
(but not used by the learning algorithm)
then the predictive accuracy of the model can be evaluated.
Different models can therefore be directly compared.

In the usual case, the training data is ``real'', 
i.e. the model is learned directly from labelled examples
that were collected specifically for that purpose.
However, quite frequently, 
modifications are made to the training data
after it is collected.
For example, it is standard practice
to remove outlier examples 
and normalise numeric values.
Moreover, the machine learning algorithm itself 
may specify modifications to the training data.
For example, the widely known bagging algorithm \citep{breiman96}
repeatedly resamples the training data with replacement, 
with each resample being used to train one of multiple models.
Similarly the random forest algorithm \citep{breiman01}
repeatedly ``projects'' the training data
in a random fashion 
in order to produce diverse decision trees.
Data augmentation strategies in which artificial data
is transformed version of the original data also falls into this category.

In our own previous work \citep{mayo14}
we introduced the idea of creating \emph{entirely new}
training data for a predictive model using an evolutionary algorithm.
In that paper, we performed a very preliminary 
evaluation in the context of classification.
In this paper, we focus 
specifically on the problem of regression 
using the naive Bayes for regression (NBR) model \citep{frank00}
and perform a much stronger evaluation.

The primary motivation for this line of research 
is to find methods to improve interpretable machine learning models.
Interpretable models are simple models easily understood by
users wanting to know, for example, why certain predictions are made,
or what knowledge the model has uncovered from the data.
A well-known drawback of such models, however, is that 
that they often lack accuracy to non-interpretable models
(e.g. deep neural networks). 
Can we therefore improve their accuracy somehow?

One approach to solving this problem is to create better interpretable
learning algorithms.
This is certainly a valid line of inquiry.
A second approach, mentioned above, 
is to preprocess the data somehow to improve the model.
How to do that most effectively is also an open question.

The third approach, and the one we focus on here,
is to keep the model and its training algorithm,
along with the training data, fixed,
and instead manipulate the inputs to the learning algorithm. 
The real training data can be used to measure the quality
of the results.
In other words,
the basic idea is to treat 
the confluence of the model, its training algorithm,
and its training data as a black box
and then try to optimise the entire system
by treating the inputs to the system (i.e., some training data)
as ``knobs" that can be tweaked to obtain new behaviours.

An advantage of this approach is that no new algorithms
need to be created: instead, existing algorithms and models can be wrapped
inside the optimiser and, as we shall see,
the potential exists for the models to actually
improve in terms of accuracy.

In order to create the artificial training data in an efficient way,
the recently proposed 
competitive swarm optimisation (CSO) algorithm \citep{cheng15}
is applied.
This algorithm optimises a fixed-length numeric vector -- a ``particle'' --
such that a given fitness function is minimised.
In our approach, 
particles correspond to artificial training datasets
and the fitness function is the error 
that a NBR model achieves after being trained on the artificial data
and tested against the training data.

When the optimisation process is complete,
the resulting NBR model
-- trained on the best artificial dataset --
is tested accordingly against held-out test data.
We also evaluate an NBR model 
trained directly on the real training data
(i.e., using the ``normal'' machine learning process)
and evaluate it against the same test data.

Our results show that for many benchmark machine learning problems,
training an NBR model indirectly on optimised artificial training data
significantly improves generalisation performance compared to direct training.
Furthermore, the artificial training datasets themselves
can be quite small, e.g., containing only three or ten examples,
compared to the original training datasets, 
which may have hundreds or thousands of examples.

The remainder of this paper is structured as follows.
In the next section, we briefly survey the two main algorithms
used in the research, specifically NBR and CSO.
We also briefly touch on previous work in artificial data generation.
In the section following that, 
we describe our new approach more formally,
and then we 
we present results from two rounds of experimentation.
The final sections of the paper
concern firstly an analysis of an artificial dataset
produced by our approach on a small 2D problem,
followed by a conclusion.

\section{Background}\label{section:background}

\subsection{Naive Bayes for Regression}
Naive Bayes is a popular approach to classification 
because it is fast and often surprisingly accurate.
The naive Bayes approach employs Bayes' rule 
by assuming conditional independence.
Given a set of $m$ categorical random variables $\{X_1, X_2, ..., X_m\}$
and another categorical random variable $Y$ whose value we want to predict,
Bayes' rule is given by
\begin{equation*}
\begin{aligned}
&P(Y = y | X_1 = x_1, X_2 = x_2, ..., X_m = x_m) = \\
&\frac{P(X_1 = x_1, X_2 = x_2, ..., X_m = x_m| Y = y) P(Y = y)}{\sum_y P(X_1 = x_1, X_2 = x_2, ..., X_m = x_m| Y = y)P(Y = y)}
\end{aligned}
\end{equation*}
Assuming that the $X_i$ are conditionally independent given $Y$, this yields the naive Bayes model:
\begin{equation*}
\begin{aligned}
&P(Y = y | X_1 = x_1, X_2 = x_2, ..., X_m = x_m) = \\
&\frac{\prod_{i=1}^m P(X_i = x_i | Y = y) P(Y = y)}{\sum_y \prod_{i=1}^m P(X_i = x_i | Y = y)P(Y = y)}
\end{aligned}
\end{equation*}
With categorical data,
it is trivial to estimate the conditional probabilities $P(X_i = x_i| Y = y)$.
The prior class probability $P(Y = y)$ is based on relative frequencies of corresponding events in the training data. 
If a predictor variable $X_i$ is continuous rather than categorical, it is common to discretise it. 
Alternatively, one can compute a normal density estimate or a kernel density estimate for $X_i$ for each different value of the categorical target variable $Y$. 

The assumption of conditional independence is rarely correct in practical applications of this model on real-world data. 
However, accurate classification does not necessarily require accurate posterior probability estimates $P(Y | X_1, X_2, ..., X_m)$:
to achieve high classification accuracy, 
it is sufficient for the correct class value $Y$ to receive maximum probability;
the absolute value of this probability is irrelevant as long as it is larger 
than the probability for any of the other (incorrect) class values.

In this paper, we consider the NBR algorithm~\citep{frank00}, 
which applies to situations with a continuous target variable $Y$.
The basic model in the regression case remains the same as the classification
case described above.
However, estimation of the required conditional densities $f(X_i = x_i|Y = y)$,
as opposed to the conditional probabilities $P(X_i = x_i| Y = y)$, is more challenging.
We provide a brief explanation here. Further details can be found in~\citep{frank00}.

Assume that $X_i$ as well as $Y$ is continuous.
We can apply the definition of conditional density to yield
\begin{equation*}
f(X_i = x_i| Y = y) = \frac{f(X_i = x_i, Y = y)}{f(Y = y)}
\end{equation*}
Both $f(X_i = x_i, Y = y)$ and $f(Y = y)$ can then be estimated using a kernel density estimator. 
In~\cite{frank00}, a Gaussian kernel is used in the kernel density estimates and the bandwidths of the kernels are estimated by minimising leave-one-out-cross-validated cross-entropy on a grid of possible bandwidth values. 
The bandwidths for $X_i$ and $Y$ are optimised jointly for $f(X_i = x_i, Y = y)$.
The obtained bandwidth for $Y$ is used to estimate both $f(X_i = x_i, Y = y)$ and $f(Y = y)$.

When $X_i$ is categorical rather than continuous, we can apply Bayes' rule to obtain
\begin{equation*}
P(X_i = x_i| Y = y) = \frac{f(Y = y| X_i = x_i)P(X_i = x_i)}{\sum_{x_i} f(Y = y| X_i = x_i)P(X_j = x_i)}
\end{equation*}
Here, $f(Y = y| X_i = x_i)$ can be estimated using a different kernel density estimator for each value $x_i$ of the categorical variable $X_i$.
The bandwidth for each of these estimators $f(Y = y | X_i = x_i)$ is again obtained using leave-one-out-cross-validated cross-entropy.
The probability $P(X_i = x_i)$ can be estimated by counting events in the training set.

The ultimate goal of naive Bayes for regression 
is prediction based on the posterior density 
$f(Y = y | X_1 = x_1, X_2 = x_2, ..., X_m = x_m)$, which is given by
\begin{equation*}
f(Y = y | X_1 = x_1, X_2 = x_2, ..., X_m = x_m) = \frac{\prod_{i=1}^m q(X_i = x_i | Y = y) f(Y = y)}{\int_y \prod_{i=1}^m q(X_i = x_i | Y = y)f(Y = y)}
\end{equation*}
where 
\begin{equation*}
q(X_i = x_i | Y = y) =
\left\{
\begin{array}{ll}
  P(X_i = x_i | Y = y)  & \mbox{if } X_i \mbox{ is categorical} \\
  f(X_i = x_i | Y = y)  & \mbox{if } X_i \mbox{ is continuous}
  \end{array}
\right.
\end{equation*}
The integral in the denominator can be obtained using a discrete approximation.
In this paper, we use the mode of the density as the predicted value, found using a recursive grid search. Note that the location of the mode can be determined without computing the denominator because its value does not depend on $Y$.

\subsection{Competitive Swarm Optimisation}

Competitive swarm optimisation (CSO) 
is a recently proposed \citep{cheng15} metaheuristic optimisation algorithm
derived from particle swarm optimisation \citep[PSO;][]{kennedy01,bonyadi17}.
The PSO family of algorithms are distinguished from
other optimisation algorithms by 
(i) not requiring a gradient function,
which enables them to tackle ill-defined problems that are otherwise
challenging; and
(ii) by performing a goal-directed mutation on solutions (or ``particles''),
in which several components from the search process 
(such as various best solutions found so far)
contribute to the search procedure.
 
Similarly to traditional PSO,
CSO represents solution vectors as particles.
For each particle in CSO, the following information is stored: 
(i) a position vector $X(t)$ 
representing a potential solution
to the current problem
and (ii) a velocity vector $V(t)$
storing $X(t)$'s trajectory through the solution space.
Parameter $t$ is the iteration index and corresponds to time.
Both algorithms maintain a single set of particles
$P(t)$ comprising $s$ particles in memory. 
This set is called the swarm.

CSO varies from traditional PSO in several important ways.
Firstly, CSO does not maintain a ``personal best'' memory per particle.
Nor does it maintain an explicit ``global best'' memory for the entire swarm.
Instead, the algorithm takes an elitist approach and runs a series
of $\sfrac{s}{2}$ fitness competitions between random pairs of particles
in the swarm.
Whichever particle in each competition
has the best fitness wins the competition.
Winners are \emph{not} updated for the next iteration.
Conversely, the losing particle's velocity and position \emph{is} updated.
Since the global best particle will never lose a fitness competition
(at least until it is improved upon),
the algorithm will always maintain the global best particle in $P(t)$.

A second key distinction between CSO and PSO
is its convergence behaviour.
Standard PSO algorithms potentially suffer
from unbounded particle velocity increases, i.e.,
particles may travel off ``to infinity'' along one or more dimensions
and therefore the algorithm will not reach an equilibrium.
Practical workarounds for this problem are either to limit particle velocities
to some maximum magnitude
or to carefully select PSO's parameters 
such that the algorithm converges \citep{bonyadi17}.
CSO requires neither of these workarounds:
\cite{cheng15} proved that as long as the key parameter $\varphi$
is non-negative,
then the particles will eventually converge on some solution
$y^*$, which may not be the global optimum
but is none-the-less
an equilibrium state for the algorithm.

\begin{algorithm} 
  initialize  $P(0)$  with $s$ particles, 
    each particle having a random initial position vector $X_{i}(0)$
    and a zero initial velocity vector $V_{i}(0)$\;                     \label{code:init} 
  $t= 0$\;
  \While{$t<t_{max}$}{\label{code:outer_loop}
     calculate the fitness of new or updated particles in $P(t)$\;
     $ U = P(t), P(t + 1) = \phi$ \;
     \While{$U \neq \phi$}{\label{code:inner_loop}
       randomly remove two particles $X_{1}(t)$, $ X_{2}(t)$ from U\;
       calculate the cost function $f(X(t))$ for $X_{1}(t)$ and $X_{2}(t)$\;\label{code:fitness_calc}
       \eIf{$f(X_{1}(t)) \leq f(X_{2}(t))$}{              \label{code:fitness_competition}
          $X_{w}(t) = X_{1}(t), X_{l}(t) = X_{2}(t);$
       }{
          $X_{w}(t) = X_{2}(t), X_{l}(t) = X_{1}(t);$
       }
       set $X_{w}(t + 1)=X_{w}(t)$ 
         and add $X_{w}(t + 1)$ to $P(t + 1)$\;              \label{code:winner_preserved}
       create three random vectors, $R_{1}(t)$, $R_{2}(t)$ and $R_{3}(t)$, 
         with elements drawn uniformly from $[0,1]$\;          \label{code:random_vectors}
       set $V_{l}(t + 1)=R_{1}(t)V_{l}(t) + R_{2}(t)(X_{w}(t)-X_{l}(t))
         +\varphi R_{3}(t)(\bar X(t)-X_{l}(t))$\;             \label{code:velocity_update}
       set $X_{l}(t + 1) =X_{l}(t) + V_{l}(t + 1)$\;          \label{code:position_update}
       add $X_{l}(t + 1)$ to $P(t + 1)$\;
	 }
  $ t=t+1$;
}
\caption{Competitive Swarm Optimiser (CSO) for large scale optimisation.}
\label{algorithm:CSO}
\end{algorithm}

Complete pseudocode for CSO, adapted from \cite{cheng15},
is given in Algorithm \ref{algorithm:CSO}.
The pseudocode assumes that a fitness function $f$
over particles, to be minimised, is provided.
Essentially, CSO consists of two loops:
an outer loop over iterations defined at line \ref{code:outer_loop},
and an inner loop over fitness competitions per iteration
defined at line \ref{code:inner_loop}.
The fitness values are computed
on line \ref{code:fitness_calc}.
Line \ref{code:winner_preserved} shows how the winner
of the current competition passes through to the next iteration unchanged.

The rules for updating the velocity and position 
of each losing particle are specified in
lines \ref{code:random_vectors}--\ref{code:position_update}.
The first of these lines creates three random vectors.
These are used to mix three different components
together in order to compute a new velocity 
for the particle being updated.
The three components are
(i) the particle's original velocity
which provides an element of inertia to the velocity updates;
(ii) the difference between the particle's current position
and the position of the particle that it just lost to;
and (iii) the difference between the particle and the swarm mean $\bar X(t)$.
The updated velocity is then added 
to the losing particle's position
so that the particle moves to a new position in search space.

A key aspect of the algorithm is
that the influence of the swarm mean is controlled by the parameter $\varphi$.
When $\varphi=0$, the swarm mean does not need to be computed 
and is effectively ignored;
when $\varphi>0$, however, 
the swarm mean has an increasingly attractive effect 
on the particles.
Overall, the algorithm has fewer parameters than other similar algorithms.
Besides $\varphi$, the only other critical parameters are
$s$, the number of particles in the swarm,
and $t_{max}$, the number of iterations to perform.
In the general case, $t_{max}$ 
can be changed to an alternative stopping criterion.

Prior evaluations \citep{cheng15}
have shown that CSO outperforms
other population-based metaheuristic algorithms 
on standard high-dimensional 
benchmark optimisation functions.
Therefore, due to its simplicity and excellent performance,
we adopt CSO for this research.

\subsection{Prior Work on Artificial Example Generation}

The creation of artificial training data 
for testing new machine learning algorithms
(as opposed to training the models)
has a long history.
The basic idea is to create training data
that has known patterns or properties,
and then observe the algorithm's performance.
For example, does the algorithm correctly detect
known patterns that exist in the artificial data?
Answering these questions can lead to insight about
how well the algorithm will perform in with real datasets.
A well-known example of an artificial dataset generator
is the ``people'' data generator described by \cite{agrawal93}.

Another class of artificial training data generation
is used for speeding up nearest neighbour classifiers.
Nearest neighbour classification can be computationally
expensive if the number and/or dimensionality
of the examples is high.
A suite of algorithms 
for optimising the set of possible neighbours
therefore exists in the literature:
see \cite{triguero11} for a comprehensive survey.
In most cases these algorithms focus on example selection
from the real data 
rather than the generation of new examples.
However, algorithms have been proposed for both modifying existing examples
(e.g., via compression as performed by \cite{zhong17}) 
and also for creating entirely new examples, 
such as in the work by \cite{impedovo14} 
which describes a method for digit prototype generation.
\cite{escalante13} also describe a method of example generation
in which genetic programming is applied.

Artificial data generation techniques 
can be used to address
the class imbalance problem.
Many machine learning algorithms for classification
fail to work effectively 
when the classes are severely imbalanced.
In the classic machine learning literature,
\cite{chawla02} developed the synthetic minority over-sampling technique (SMOTE)
to artificially increase the size of the minority class,
(producing better class balance in the data)
while correspondingly 
preserving the examples in the majority class.
Similarly,
\cite{lopez14} developed a new technique based on evolutionary
algorithms for generating minority class examples. 

It is worth mentioning here the significant number
of works in the literature showing that corrupting training
data with noise 
improves the resulting generalisation performance of models.
This idea is common in the neural network literature.
For example, \cite{an96} found that input
data corruption improved neural network generalisation for both
classification and regression tasks.
More recently, stacked denoising autoencoders \citep{vincent08},
which are based on the idea of corrupting inputs 
to learn more robust features in deep networks,
have seen considerable success.
Likewise, in the evolutionary machine learning literature,
dataset corruption has been demonstrated to be an effective
strategy for learning both classifier systems \citep{urbanowicz11}
and symbolic discriminant analysis models \citep{greene09}.
These approaches do not strictly generate artificial examples;
rather they are based on perturbing the real training data.

Finally, we briefly mention
generative adversarial nets (GANs) proposed
by \cite{goodfellow14}.
These are a type of paired neural network
in which one network ``generates'' artificial data examples
by passing random noise through the network,
and the other network ``discriminates'' between the examples
in order to 
to determine if the examples are from the real training dataset
or are fakes.
The purpose of this process is to construct better generative
models from the data.

In contrast,
the work we present in this paper extends and build on
work originally reported in \cite{mayo14}.
Rather than aiming to generate fake data that is indistinguishable
from real data, we aim to produce artificial data 
maximising the performance of the resulting classifier
when it is trained on the artificial data.
Therefore, the artificial data may (or may not)
be easily distinguishable from the real data.
In fact, in some cases the artificial data is easily
distinguishable from the real data.
For example, an artificial dataset depicted discussed shortly
has negative values for the prediction target:
such values do not occur in the corresponding real data. 
The approach proposed in this paper is
therefore quite different in nature to the GAN approach.

To summarise, prior work has explored
many variations on the theme of perturbing the
training data in different ways in order to improve
model performance, or focussed on example generation
strategies for specific purposes (such as balancing the data).
\cite{mayo14}
is the only prior work investigating
the generation of completely new artificial training datasets.
In that work, naive Bayes for classification was shown to
perform well with the optimised artificial data.
In this paper, we significantly extend that work by 
(i) applying the same basic approach to naive Bayes for regression;
(ii) updating the metaheuristic algorithm used for the optimisation
to CSO, a more modern algorithm designed specifically for high dimensional
problems; and
(iii) testing against a wide range of varied regression problems.

\section{Evolving Artificial Data for Regression}\label{section:approach}

In this section, we precisely outline
our approach to constructing artificial training data
for regression problems using CSO.

The classical approach to machine learning 
is to start with some tabular training data $D_{train}$,
comprising $r$ rows or examples,
and $d$ attributes.
One of the attributes, typically the first or last one, 
is the target to predict and the other attributes
are predictor variables.

Given this training data, we can think of a machine learning
algorithm such as NBR as a function of the training data,
i.e.,
\begin{equation*}
m_1=m_{NBR}(D_{train})
\end{equation*} 
where $m_1$ is a predictive model
and $m_{NBR}$ is a function encapsulating a machine learning algorithm
that trains an NBR model given some training dataset.
The performance of the model on a test dataset
can similarly be stated as a function
\begin{equation*}
e_1=eval(m_1,D_{test})
\end{equation*}
with $e_1$ being a standard metric for regression error 
such as root mean squared error (RMSE).

Our approach is based on
defining a corresponding fitness function $f(X(t))$ for CSO.
In order to evaluate $f(X(t))$,
we must first convert particle $X(t)$ into a tabular dataset
with $n$ examples
where $n$ is small and fixed.
This we will refer to as $D_{X(t)}$. 
Typically we will have $n\ll r$ 
unless $r$ itself is very small.
Since the entire dataset $D_{X(t)}$ has $dn$ unique values,
then the dimensionality of the particles $X(t)$ is $dn$.

The process of converting $X(t)$ to $D_{X(t)}$ is fairly straightforward:
for numeric attributes, the values are copied directly from the particle
into the correct position in $D_{X(t)}$ without modification.
For categorical (i.e., discrete) attributes, the process
is complicated by the fact that CSO optimises continuous values only.
We correct for this by rounding individual values from $X(t)$ 
to the nearest legal discrete value 
when they are copied across to $D_{X(t)}$.
The original values in $X(t)$ remain unmodified.

Random particle initialisation 
(line \ref{code:init} of Algorithm \ref{algorithm:CSO}),
is performed by an examination of $D_{train}$:
each variable is initialised to a random uniform value
between the minimum and maximum values of the corresponding
attribute in $D_{train}$.

Thus, we can define the fitness function to minimise as
the error some model $m_2$ yields on the training data:
\begin{equation*}
f(X(t))=eval(m_2,D_{train})
\end{equation*}
\noindent where $m_2$ is not trained on the original
training data, but instead on the artificial dataset
that is derived from the current particle $X(t)$ being evaluated,
i.e.
\begin{equation*}
m_2=m_{NBR}(g(X(t)))=m_{NBR}(D_{X(t)})
\end{equation*}
where the function $g$ converts the one-dimensional particle
$X(t)$ into a tabular dataset $D_{X(t)}$
using the procedure outlined above.

After the optimisation process is complete,
we can evaluate $m_2$ against the test data
to get the test error 
\begin{equation*}
e_2=eval(m_2,D_{test})
\end{equation*}
This will enable us to perform a direct comparison with $e_1$,
the test error achieved using the traditional machine learning approach.

In terms of computational time complexity,
the significant cost of this algorithm is the fitness function
evaluation.
Evaluating the fitness of a particle
involves firstly converting the particle
to a dataset;
secondly, training a regression model using the dataset;
and thirdly,
evaluating that model against the real dataset.
The training and testing steps clearly have the most significant complexity
in this operation.
Since CSO needs at least a few hundred iterations to find good solutions,
this implies that our approach may not be feasible
with models that are very time consuming to train or test,
which may be due to either the specific learning algorithm's own complexity
or the real dataset being very large in size.
Measures may therefore need to be taken 
(e.g. replacing the algorithm with a more efficient
one, or only using a subset of the real data for fitness evaluations)
in these cases.

As mentioned, we have chosen a direct representation in which
variables being optimised map directly to attribute values in the artificial
dataset.
A consequence of this is that different particles (the equivalent
of ``genotypes'' in evolutionary computing)
may result in identical models (i.e., the equivalent of ``phenotypes'')
if the learning algorithm ignores the order of the examples
in the data.
For example, the datasets $\{x_1,x_2,x_3\}$ and $\{x_2,x_3,x_1\}$,
where $x_i$ is an example,
produce the same NBR model.
This observation is not necessarily a drawback of our approach.
In fact, in many other approaches,
multiple genotypes mapping 
onto a single phenotype is a deliberate feature
as it enables neutral mutations.
For example, linear genetic programming (LGP),
proposed by \cite{brameier07},
evolves sequential programs. 
Since the order of any two instructions
is often not dependent (e.g., initialising
two different variables),
the same issue arises:
different programs map to the same behaviours.
However, this does suggest
that future research could investigate alternative
particle/dataset mapping functions to determine if
efficiencies are possible over the approach
we use here.

\section{Initial Evaluation}\label{section:evaluation-small}

We conducted an initial evaluation 
of our proposed algorithm
on benchmark regression datasets used in 
\cite{frank00}
\footnote{Original datasets available from 
\url{http://prdownloads.sourceforge.net/weka/datasets-numeric.jar}}.
Although small,
these datasets do represent a diverse range
of prediction problems and have been acquired
from several different sources.

To illustrate the diversity,
the ``fruit fly'' dataset consists of observations
describing male fruit fly sexual behaviour
and is used to build a model predicting fly longevity.
Conversely, the ``cpu'' dataset is used to construct a predictive model
for CPU performance given various technical features of a CPU.
Finally, the ``pollution'' dataset is used to predict the age-adjusted
mortality rate of a neighbourhood given various metrics
such as the amount of rainfall and the degree of presence
of various pollutants.
Overall, there are 35 distinct regression datasets in the collection,
with between 38 to 2178 examples per dataset.
The dimensionality of the datasets ranges from 2 to 25. 
Table \ref{table:dataset-characteristics} gives the exact properties of each dataset.

\begin{table}
\caption{Dataset characteristics used in the experiments.
The third column shows the total number of features,
with the total number of numeric features in parentheses.}
\small
\rowcolors{2}{gray!25}{white}
\begin{tabular}{lrr}
\rowcolor{gray!50}
\hline
Dataset & \#Examples & \#Features\\ 
\hline
auto93 & 93 & 22(16)\\
autoHorse & 205 & 25(17)\\
autoMpg & 398 & 7(4)\\
autoPrice & 159 & 15(15)\\
baskball & 96 & 4(4)\\
bodyfat & 252 & 14(14)\\
bolts & 40 & 7(7)\\
cholesterol & 303 & 13(6)\\
cleveland & 303 & 13(6)\\
cloud & 108 & 6(4)\\
cpu & 209 & 7(6)\\
detroit & 13 & 13(13)\\
echoMonths & 130 & 9(6)\\
elusage & 55 & 2(1)\\
fishcatch & 158 & 7(5)\\
fruitfly & 125 & 4(2)\\
gascons & 27 & 4(4)\\
housing & 506 & 13(12)\\
hungarian & 294 & 13(6)\\
longley & 16 & 6(6)\\
lowbwt & 189 & 9(2)\\
mbagrade & 61 & 2(1)\\
meta & 528 & 21(19)\\
pbc & 418 & 18(10)\\
pharynx & 195 & 11(1)\\
pollution & 60 & 15(15)\\
pwLinear & 200 & 10(10)\\
quake & 2178 & 3(3)\\
schlvote & 38 & 5(4)\\
sensory & 576 & 11(0)\\
servo & 167 & 4(0)\\
sleep & 62 & 7(7)\\
strike & 625 & 6(5)\\
veteran & 137 & 7(3)\\
vineyard & 52 & 3(3)\\
\hline
\end{tabular}
\label{table:dataset-characteristics}
\end{table}

Certainly, the datasets themselves 
are not ``high dimensional'' datasets by any standard.
However, when the proposed optimisation algorithm 
is applied to evolve artificial datasets 
with, e.g., ten examples,
then the dimensionality for the optimisation problem
ranges between 30 and 260 depending on the dataset.
Thus, the dimensionality of the optimisation problem high-dimensional.

To prepare the datasets for our evaluation,
we split each dataset into a training portion
(consisting of a randomly selected 66\% of the examples)
and a held-out test portion
(consisting of the remaining examples).
Maintaining a fixed test set was necessary 
so that the variability in the results produced by 
our stochastic search method could be assessed properly.

The key parameters for CSO, specifically $\varphi$, $t_{max}$ and $s$
were set of 0.1, 1000, and 100 respectively.
For evaluating the models, the fitness function to be minimised
was RMSE,
and we fixed the size of the artificial datasets to $n=10$.
These initial parameter values were chosen after some initial exploratory
tests on one of the datasets.

Because CSO is a heavily randomised search algorithm 
and will therefore give different results
each time it is run,
we ran our algorithm ten times for on each train/test split.
Thus, we executed a total of 350 runs of our algorithm overall.

\begin{table}\caption{Initial experiment results on the small datasets.
Shown are the test errors.
To enhance readability, an asterisk ($^{*}$) indicates the 
best result in an LR vs. NBR vs. CSO-NBR(mean) comparison,
while \textbf{bold} indicates the best result in a NBR vs. CSO-NBR(mean) comparison.
We also include the results achived by parameter-optimised GPR as a reference.}
\small
\rowcolors{2}{gray!25}{white}
\begin{tabular}{lrrrrrr}
\rowcolor{gray!50}
\hline
Dataset & GPR & LR & NBR & CSO-NBR(mean) & $p$ value & CSO-NBR(best)\\ 
\hline 
     auto93 &5.00&       11.08 &     $^{*}$\bf5.41 &        5.91 $\pm$        1.89 &         0.7880 &       3.92 \\
  autoHorse &11.19&       13.87 &    $^{*}$\bf12.92 &       13.00 $\pm$        2.59 &         0.5401 &       9.26 \\
    autoMpg &3.02&     $^{*}$3.15 &        \bf3.43 &        3.55 $\pm$        0.12 &         0.9926 &       3.32 \\
  autoPrice &2503.06&     2640.96 &     2375.45 &  $^{*}$\bf2261.03 $\pm$      218.19 &         0.0658 &    1894.58 \\
   baskball &0.08&     $^{*}$0.08 &        0.11 &        \bf0.09 $\pm$        0.01 &    0.0007 &       0.08 \\
    bodyfat &0.64&        0.72 &        1.68 &     $^{*}$\bf0.53 $\pm$        0.09 &    0.0000 &       0.41 \\
      bolts &13.93&       11.29 &     $^{*}$\bf7.92 &       11.46 $\pm$        8.35 &         0.8938 &       7.52 \\
cholesterol &51.17&    $^{*}$50.95 &       61.76 &       \bf56.92 $\pm$        3.81 &   0.0015 &      51.63 \\
  cleveland &0.77&     $^{*}$0.79 &        1.05 &        \bf0.80 $\pm$        0.04 &    0.0000 &       0.74 \\
      cloud &0.32&     $^{*}$0.31 &        0.43 &        \bf0.37 $\pm$        0.12 &         0.0881 &       0.31 \\
        cpu &31.68&       93.75 &      114.65 &    $^{*}$\bf43.70 $\pm$        6.85 &    0.0000 &      28.37 \\
    detroit &27.8&     $^{*}$48.24 &       89.18 &       \bf52.24 $\pm$       20.06 &    0.0001 &      21.79 \\
 echoMonths &11.22&     $^{*}$11.21 &       \bf12.67 &       12.89 $\pm$        1.70 &         0.6526 &      11.32 \\
    elusage &14.81&     $^{*}$12.94 &       17.73 &       \bf17.52 $\pm$        1.26 &         0.3002 &      16.12 \\
  fishcatch &61.61&      104.76 &      209.45 &     $^{*}$\bf63.68 $\pm$       18.68 &    0.0000 &      45.83 \\
   fruitfly &14.94&     $^{*}$14.32 &       19.80 &       \bf16.94 $\pm$        0.70 &    0.0000 &      15.67 \\
    gascons &11.18&       40.27 &       17.63 &      $^{*}$\bf3.23 $\pm$        0.76 &    0.0000 &       2.17 \\
    housing &3.64&        5.20 &        6.32 &      $^{*}$\bf3.92 $\pm$        0.14 &    0.0000 &       3.70 \\
  hungarian &0.31&     $^{*}$0.30 &        0.36 &        \bf0.33 $\pm$        0.02 &    0.0001 &       0.30 \\
    longley &425.42&      803.03 &   $^{*}$\bf552.15 &      755.90 $\pm$      428.00 &         0.9168 &     282.72 \\
     lowbwt &491.03&    $^{*}$496.61 &      559.99 &      \bf550.26 $\pm$       47.09 &         0.2649 &     501.34 \\
   mbagrade &0.30&      $^{*}$0.30 &      $^{*}$\bf0.30 &        0.34 $\pm$        0.02 &         1.0000 &       0.33 \\
       meta &321.00&      398.92 &      378.52 &   $^{*}$\bf365.57 $\pm$      121.54 &         0.3719 &     217.87 \\
        pbc &947.50&   $^{*}$970.26 &     1049.74 &     \bf1016.99 $\pm$       49.78 &    0.0336 &     922.27 \\
    pharynx &288.27&      429.51 &      354.80 &   $^{*}$\bf327.16 $\pm$       15.86 &   0.0002 &     301.58 \\
  pollution &46.36&     $^{*}$46.29 &       62.31 &       \bf55.84 $\pm$        8.78 &    0.0224 &      42.64 \\
   pwLinear &1.85&        2.45 &        2.49 &     $^{*}$\bf2.04 $\pm$        0.05 &    0.0000 &       1.96 \\
      quake &0.20&        0.20 &        0.27 &     $^{*}$\bf0.19 $\pm$        0.00 &    0.0000 &       0.19 \\
   schlvote &1855748.32&  1795017.06 &  1956688.47 &$^{*}$\bf1096037.48 $\pm$    67480.04 &    0.0000 & 1023308.85 \\
    sensory &0.75&     $^{*}$0.79 &        0.94 &        \bf0.82 $\pm$        0.02 &    0.0000 &       0.79 \\
      servo &0.74&     $^{*}$0.84 &        1.04 &        \bf0.88 $\pm$        0.05 &    0.0000 &       0.82 \\
      sleep &2.77&     $^{*}$2.79 &        4.09 &        \bf3.49 $\pm$        0.55 &    0.0038 &       2.81 \\
     strike &657.92&   $^{*}$662.13 &      696.60 &      \bf680.04 $\pm$       29.17 &         0.0531 &     663.30 \\
    veteran &157.83&   $^{*}$155.44 &      \bf164.17 &      181.64 $\pm$       31.27 &         0.9444 &     155.39 \\
   vineyard &2.52&        1.80 &     $^{*}$\bf1.33 &        2.72 $\pm$        0.56 &         1.0000 &       2.16 \\
\hline
\end{tabular}
\label{table:initial-experiment}
\end{table}

Table \ref{table:initial-experiment} gives the results by dataset.
The column ``NBR'' in the table
gives the baseline testing performance of NBR trained
using the traditional process.
We further include in the table the results achieved
using linear regression (LR) and Gaussian process regression with radial basis function kernels (GPR) as implemented in WEKA 3.8.0 \citep{frank16}.
The implementation of LR
performs attribute selection using the M5$^\prime$ 
method \citep{wang97}, which usually outperforms standard LR.
We tuned the kernel parameter and the noise parameter of GPR using a grid search that optimised parameters based on internal cross-validation of the root mean squared error on the training set. We used the range $10^0...10^{-5}$ for the noise parameter and the range $10^5...10^{-5}$ for the kernel parameter.
Note that GPR is a state-of-the-art black box regression method and therefore
it is expected that it will outperform methods that produce simple models.

The error metrics on the test splits
achieved by our experimental algorithm 
are given in the columns ``CSO-NBR(mean)'' and ``CSO-NBR(best)'',
which give the average ($\pm$ the standard deviation)
and best errors achieved respectively over the ten repetitions.

Finally, we also performed a single-sample, one-sided $t$-test
comparing the results of our experimental algorithm
with the single result achieved by traditional NBR.
The $p$ value for the test 
is given as well in Table \ref{table:initial-experiment}.

Overall, we can see NBR accuracy performance
is improved significantly in most cases
when CSO-NBR is used to train the model.
If we consider the strong condition that $p<0.01$ (i.e., 99\% confidence),
then NBR's performance is improved in $\sfrac{18}{35}$ cases.
If the condition is weakened to $p<0.05$ (95\% confidence),
then a further two more significant improvements
can be added to that total.

Also interesting is the comparison of NBR, LR,
and our experimental algorithm.
NBR in its original form tends to underperform LR
on most of the problems;
in fact, on $\sfrac{25}{35}$ LR has a lower error
than NBR.
However, when we compare the mean performance of CSO-NBR
to LR, the number of wins for LR decreases to $\sfrac{21}{35}$.
If we furthermore consider
the error of the best single run 
of our experimental algorithm, 
then the number of LR wins 
drops to only $\sfrac{8}{35}$ datasets.

We note that considering the best-of-ten-runs
result (i.e., the ``CSO-NBR(best)'' column)
is an optimistic strategy,
and it has consequently been de-emphasized 
in Table \ref{table:initial-experiment}. 
However, it is worth including these results
because in practice, if only one model is required
but CSO-NBR is run multiple times (which it should be),
then only the best single model during testing is likely to
be used for subsequent analysis 
(e.g., for inspecting model, or for use in live production).

In terms of individual dataset performance,
the difference between NBR and CSO-NBR is quite variable.
For example, the ``gascons'' dataset
appears to be very difficult to model 
for both NBR and LR, 
which achieve test set errors of 17.63 and 40.27 respectively.
CSO-NBR, on the other hand,
generalises quite well,
achieving a mean testing error of only 3.23.
A similar example is the ``bodyfat'' problem,
in which test errors of 1.68 and 0.72 for NBR and LR respectively
can be compared to the 0.53 mean test error
that CSO-NBR achieves.

Conversely, there are also datasets where CSO-NBR
performs extremely poorly: 
for example, the ``longley'' dataset is modelled well
by NBR but poorly by both LR and CSO-NBR.
Similarly for the ``vineyard'' dataset.
There appears to be no clear-cut single best algorithm overall.

Also interesting is a pattern 
that is clearly evident in the table 
of results.
If one considers the cases where CSO-NBR
is statistically significantly better than NBR,
it appears to also be the case (in all but one instance)
that LR also shows a large reduction in error
compared to NBR.
This suggests that those datasets have properties
that NBR cannot model well by itself,
without being embedded inside CSO.
This certainly warrants future investigation, perhaps in the context
of using synthetic data to determine which patterns NBR and CSO-NBR
can and cannot detect.

In terms of a comparison between the methods that produce
simple and interpretable models and the parameter-optimised
black box method GPR, as expected, GPR achieves
significant performance improvements on average.
However, in five cases (namely ``autoPrice", ``bodyfat'',``gascons'', ``quake''
and ''schlvote"), 
CSO-NBR outperforms GBR, sometimes by a wide margin.

Finally, we were also interested in a comparison of CSO
to a more traditional PSO variant.
This was to determine if CSO
was really a suitable optimiser for this problem.
To that end, we selected the standard PSO algorithm (SPSO)
described by \cite{bonyadi17}.
SPSO has three main parameters:
an inertia parameter, a cognitive constant,
and a social constant.
Following the recommendation by \cite[][pp. 11]{bonyadi17}
we set these parameters to 0.6, 1.7 and 1.7 respectively. 

A fundamental difference between the CSO and SPSO 
is that SPSO updates \emph{all} particles in the swarm
at each iteration whereas CSO updates only \emph{half}
of the particles (the losers) per iteration.
Similarly, particles in CSO have no personal memory;
particles in SPSO on the other hand 
each have a memory for their
``personal best'' position additional to their current position.
Therefore it is unfair to directly compare SPSO
and CSO with the same swarm size and same number of iterations,
as SPSO would be performing approximately twice the number 
of function evaluations, and furthermore,
it would use twice the amount of memory as CSO.

To make comparisons fairer, we gave
each algorithm 
approximately the same number of fitness function evaluations,
since the fitness function is the main contributor 
to the algorithm complexity.
There are two ways of making the number of fitness evaluations 
performed by SPSO approximately equal to those performed by CSO:
halving the swarm size or halving the number of iterations.
We tested both variants, i.e., 
we defined one version of SPSO-NBR 
with a $s=50$ and $t_{max}=1000$,
and another with $s=100$ and $t_{max}=500$.
The latter version of SPSO
uses the twice the amount of memory
as the other two algorithms.

The SPSO results are given 
in the tables
in the appendix.
A comparison of these tables
with the CSO-NBR results in
Table \ref{table:initial-experiment} 
show that in most cases,
CSO-NBR outperforms SPSO-NBR.
Specifically, the $s=50,t_{max}=1000$ variant of SPSO-NBR
outperforms CSO-NBR on only six of the datasets
while the $s=100,t_{max}=500$ variant (using twice as much memory)
outperforms CSO-NBR on twelve of the datasets.

We continue in the next section
using CSO only, as (i) it has been shown that CSO generally outperforms SPSO
on the same problems, or is at least in the same ballpark; 
and (ii) CSO requires less parameter tuning compared to SPSO.
To summarise, our initial experiments 
indicate that our experimental algorithm
-- training an NBR model on artificial training data 
optimised in a supervised fashion using CSO --
is an effective approach for constructing more robust naive Bayes
regression models for certain machine learning problems.
We therefore proceed to a more challenging set of problems in
the next round of experiments.

\section{Parameter Study with Modern Datasets}\label{section:evaluation-large}

In this set of experimental runs,
we consider the effect of various key parameters
on the artificial data optimisation process,
in particular CSO's $\varphi$ parameter,
and $n$, the size of the artificial dataset.

We ran experiments with $\varphi \in \{0.0, 0.1, 0.5, 1.0\}$
and $n \in \{3,10,20\}$ to evaluate the effect 
on the generalisation error of the resulting NBR model.
The remaining parameters, specifically $s$, the swarm size,
and $t_{max}$, the number of iterations,
were held at constant values 
(100 and 1000 respectively) so that the total number of
function evaluations was a constant.
Recall that each function evaluation consists of
one run of NBR on the artificial data to train the model 
followed by a testing procedure
to measure the RMSE on the real training data.

We chose eight new regression problems 
for this round of experiments.
These datasets are larger than those used in the previous
round of experiments,
and also much more recently published.
Table \ref{table:datasets} summarises the datasets, 
giving their name and the overall dimensionality of the data
in terms of number of examples and number of predictive
features excluding the target value.

\begin{table}
\caption{Datasets used in the experiment.}
\small
\rowcolors{2}{gray!25}{white}
\begin{tabular}{lrr}
\rowcolor{gray!50}
\hline
Dataset & \#Examples & \#Features\\ 
\hline
Wine Quality -- Red \citep{cortez09}          &1,599&11\\
Wine Quality -- White \citep{cortez09}        &4,898&11\\
Energy Efficiency -- Heating \citep{tsanas12} &768&8\\
Energy Efficiency -- Cooling \citep{tsanas12} &768&8\\
Parkinson's -- Motor \citep{tsanas14}         &5,875&18\\
Parkinson's -- Total \citep{tsanas14}         &5,875&18\\
Appliance Energy \citep{candanedo17}          &19,373&26\\
Survival \citep{lenz08,wang15}                &181 (train)/233 (test) &3,833\\
\hline
\end{tabular}
\label{table:datasets}
\end{table}

In brief, 
the wine quality datasets \citep{cortez09}
contains data about various Portuguese ``Vinho Verde'' wines.
The features are physiochemical statistics about each wine
(e.g., the pH value)
and the target to predict is the subjective quality of the wine,
which ranges on a scale from 0 to 10.
The data is divided into two separate prediction problems,
one for red wine and one for white wine
with the white wine dataset being considerably
larger in size.

The building energy efficiency datasets \citep{tsanas12},
in contrast,
concerns predicting the heating and/or cooling load of a building, 
given various properties of the building's design.
The buildings are simulated, 
and while each building has the same volume, 
the surface areas, dimensions, and the materials
used to construct the buildings vary.
There are eight predictive features in total,
and for each example there are 
two separate prediction targets:
(i) the total heating load and (ii) the total cooling load.
Accurately predicting the loads of the buildings
without running full (and highly computationally complex)
building simulations is an important
problem in the optimisation of energy efficient buildings.

The fifth and sixth problems
deal with the remote diagnosis of Parkinson's disease \citep{tsanas14}.
The aim of this data is to learn a model that can predict the 
Unified Parkinson's Disease Rating Scale (UPDRS)
for a patient.
Normally, the UPDRS score 
is determined by trained medical personnel
in the clinic using a costly physical exam.
In contrast,
the dataset constructed for this problem
uses statistical features extracted 
via the signal processing of recorded speech alone.
Thus, UPDRS assessment can be performed
remotely instead of in-person if the predictions are accurate enough.
Similar to the previous problem, this prediction problem
has two predictive targets per example:
the total UPDRS score, ranging from 0 (healthy)
to 176 (total disability);
and the motor UPDRS score, which has a
smaller range of 0 to 108.

The seventh problem, like the building energy prediction
problems, is concerned with energy usage \citep{candanedo17}.
Unlike those problems, however, this dataset is
derived from real data rather than a simulation
and is much larger: 
there are nearly twenty thousand examples in total,
and the dimensionality is higher as well.
The problem is to predict the energy usage of appliances
in a real house given features related to the environment,
such as weather data and temperature/humidity readings
from wireless sensors.
The data is arranged into a time series
with each example corresponding to data about the weather
and household/appliance usage activity over a ten minute period.

Finally, the eighth dataset we considered
concerns building a regression model 
to predict survival time for cancer patients
given gene expression data.
The data was originally acquired by \cite{lenz08}
and subsequently further processed and re-published by \cite{wang15}.
The features in the dataset are measurements from
probe sets obtained using a microarray device.
Each measurement is an estimate
of one particular gene's
degree of expression in a patient.
The prediction target is the survival time of the patient.
Because of the high dimensionality of the data, 
we performed our experiments on this data 
using a random subset of one hundred of the gene expressions measurements.
Such an approach is commonly used elsewhere.
For example, \cite{kapur16} used the same strategy when
learning Bayesian network models
for gene expression class prediction.

In terms of the experimental setup,
our experiments followed the same basic procedure
that we applied previously:
i.e., the data was split into training and test portions,
and then both standard NBR and LR served as baselines.
We ran CSO-NBR ten times to assess
its average and best performance on each problem.

With respect to splitting the data into training and test subsets,
for the first six datasets, the ordering of the examples was randomised
and the split sizes were two-thirds and one-third
for the training and test sets respectively.
The Appliance Energy data, however, forms a time series.
Therefore we did not randomise the examples
but instead preserved their order, which resulted in
the training examples preceding the test examples in time.
The split sizes were the same as before however.
For the Survival data, 
the examples are already arranged into
training and test subsets by \cite{wang15},
and we did not alter this.

\begin{table}
\caption{Experiment results (test errors) on the Wine Quality -- Red dataset.}
\small
\rowcolors{2}{gray!25}{white}
\begin{tabular}{lrrrrr}
\rowcolor{gray!50}
\hline
$\varphi$ & $n$ & NBR & CSO-NBR(mean) & CSO-NBR(best)\\ 
\hline 
        0.0 &           3 &      0.8522 &      0.6735 $\pm$      0.0037&      0.6670 \\
        0.0 &          10 &      0.8522 &      0.6745 $\pm$      0.0079&      0.6651 \\
        0.0 &          20 &      0.8522 &      0.6885 $\pm$      0.0164&      0.6656 \\
        0.1 &           3 &      0.8522 &      0.6703 $\pm$      0.0066&      0.6613 \\
        0.1 &          10 &      0.8522 &      0.6751 $\pm$      0.0098&      0.6591 \\
        0.1 &          20 &      0.8522 &      0.6756 $\pm$      0.0072&      0.6630 \\
        0.5 &           3 &      0.8522 &      0.6757 $\pm$      0.0066&      0.6641 \\
        0.5 &          10 &      0.8522 &      0.6668 $\pm$      0.0090&      0.6543 \\
        0.5 &          20 &      0.8522 &      0.6711 $\pm$      0.0072&      0.6579 \\
        1.0 &           3 &      0.8522 &      0.6775 $\pm$      0.0090&      0.6620 \\
        1.0 &          10 &      0.8522 &      0.6705 $\pm$      0.0122&      0.6534 \\
        1.0 &          20 &      0.8522 &   \bf0.6664 $\pm$      0.0099&   \bf0.6519 \\
\hline
\end{tabular}
\label{table:experiment-red-wine}
\end{table}

\begin{table}
\caption{Experiment results (test errors) on the Wine Quality -- White dataset.}
\small
\rowcolors{2}{gray!25}{white}
\begin{tabular}{lrrrrr}
\rowcolor{gray!50}
\hline
$\varphi$ & $n$ & NBR & CSO-NBR(mean) & CSO-NBR(best)\\ 
\hline 
        0.0 &           3 &      1.2724 &      0.8270 $\pm$      0.0095&      0.8107 \\
        0.0 &          10 &      1.2724 &      0.8285 $\pm$      0.0101&      0.8144 \\
        0.0 &          20 &      1.2724 &      0.8624 $\pm$      0.0734&      0.8266 \\
        0.1 &           3 &      1.2724 &      0.8272 $\pm$      0.0105&      0.8138 \\
        0.1 &          10 &      1.2724 &      0.8202 $\pm$      0.0069&      0.8133 \\
        0.1 &          20 &      1.2724 &   \bf0.8200 $\pm$      0.0113&   \bf0.8042 \\
        0.5 &           3 &      1.2724 &      0.8319 $\pm$      0.0098&      0.8171 \\
        0.5 &          10 &      1.2724 &      0.8285 $\pm$      0.0068&      0.8202 \\
        0.5 &          20 &      1.2724 &      0.8246 $\pm$      0.0093&      0.8134 \\
        1.0 &           3 &      1.2724 &      0.8361 $\pm$      0.0094&      0.8210 \\
        1.0 &          10 &      1.2724 &      0.8256 $\pm$      0.0083&      0.8109 \\
        1.0 &          20 &      1.2724 &      0.8237 $\pm$      0.0056&      0.8152 \\\hline
\end{tabular}
\label{table:experiment-white-wine}
\end{table}

\begin{table}
\caption{Experiment results (test errors) on the Energy Efficiency -- Heating dataset.}
\small
\rowcolors{2}{gray!25}{white}
\begin{tabular}{lrrrrr}
\rowcolor{gray!50}
\hline
$\varphi$ & $n$ & NBR & CSO-NBR(mean) & CSO-NBR(best)\\ 
\hline 
        0.0 &           3 &      6.2906 &      3.9545 $\pm$      0.4241&      3.4115 \\
        0.0 &          10 &      6.2906 &      4.4217 $\pm$      0.8192&      3.1010 \\
        0.0 &          20 &      6.2906 &      4.9732 $\pm$      0.9185&      3.5728 \\
        0.1 &           3 &      6.2906 &   \bf3.7849 $\pm$      0.6211&   \bf2.5752 \\
        0.1 &          10 &      6.2906 &      3.9914 $\pm$      0.4370&      2.9801 \\
        0.1 &          20 &      6.2906 &      4.5730 $\pm$      1.6502&      2.8295 \\
        0.5 &           3 &      6.2906 &      4.3207 $\pm$      0.8469&      3.5456 \\
        0.5 &          10 &      6.2906 &      4.6647 $\pm$      0.4799&      3.6592 \\
        0.5 &          20 &      6.2906 &      4.4657 $\pm$      1.0296&      2.9383 \\
        1.0 &           3 &      6.2906 &      4.0320 $\pm$      0.5351&      3.3858 \\
        1.0 &          10 &      6.2906 &      5.0931 $\pm$      0.9932&      3.5922 \\
        1.0 &          20 &      6.2906 &      4.7041 $\pm$      0.6888&      3.8492 \\\hline
\end{tabular}
\label{table:experiment-energy-heating}
\end{table}

\begin{table}
\caption{Experiment results (test errors) on the Energy Efficiency -- Cooling dataset.}
\small
\rowcolors{2}{gray!25}{white}
\begin{tabular}{lrrrrr}
\rowcolor{gray!50}
\hline
$\varphi$ & $n$ & NBR & CSO-NBR(mean) & CSO-NBR(best)\\ 
\hline 
        0.0 &           3 &      5.5104 &      3.5554 $\pm$      0.4713&      2.8654 \\
        0.0 &          10 &      5.5104 &      4.0276 $\pm$      0.5111&      3.3790 \\
        0.0 &          20 &      5.5104 &      4.9107 $\pm$      0.4092&      4.1077 \\
        0.1 &           3 &      5.5104 &      3.5219 $\pm$      0.3873&      2.8390 \\
        0.1 &          10 &      5.5104 &   \bf3.2015 $\pm$      0.3916&   \bf2.5128 \\
        0.1 &          20 &      5.5104 &      4.4863 $\pm$      2.4956&      2.9844 \\
        0.5 &           3 &      5.5104 &      3.4576 $\pm$      0.3586&      2.9568 \\
        0.5 &          10 &      5.5104 &      3.4896 $\pm$      0.5219&      2.5689 \\
        0.5 &          20 &      5.5104 &      4.0933 $\pm$      0.8163&      3.1503 \\
        1.0 &           3 &      5.5104 &      3.4930 $\pm$      0.2802&      3.0616 \\
        1.0 &          10 &      5.5104 &      4.2263 $\pm$      1.1971&      3.1754 \\
        1.0 &          20 &      5.5104 &      3.7196 $\pm$      0.4081&      3.0461 \\\hline
\end{tabular}
\label{table:experiment-energy-cooling}
\end{table}

\begin{table}
\caption{Experiment results (test errors) on the Parkinson's -- Motor dataset.}
\small
\rowcolors{2}{gray!25}{white}
\begin{tabular}{lrrrrr}
\rowcolor{gray!50}
\hline
$\varphi$ & $n$ & NBR & CSO-NBR(mean) & CSO-NBR(best)\\ 
\hline 
        0.0 &           3 &     10.8869 &      7.8192 $\pm$      0.1168&      7.6027 \\
        0.0 &          10 &     10.8869 &      8.2723 $\pm$      0.5394&      7.4407 \\
        0.0 &          20 &     10.8869 &   \bf7.6298 $\pm$      0.8589&   \bf6.5968 \\
        0.1 &           3 &     10.8869 &      8.2869 $\pm$      0.3250&      7.8042 \\
        0.1 &          10 &     10.8869 &      8.4412 $\pm$      0.3450&      8.0629 \\
        0.1 &          20 &     10.8869 &      8.4356 $\pm$      0.4726&      7.9741 \\
        0.5 &           3 &     10.8869 &      7.9885 $\pm$      0.2380&      7.5894 \\
        0.5 &          10 &     10.8869 &      7.8277 $\pm$      0.2719&      7.2337 \\
        0.5 &          20 &     10.8869 &      8.0523 $\pm$      0.4004&      7.5780 \\
        1.0 &           3 &     10.8869 &      7.8931 $\pm$      0.3850&      7.3222 \\
        1.0 &          10 &     10.8869 &      7.7304 $\pm$      0.3464&      7.2379 \\
        1.0 &          20 &     10.8869 &      8.0553 $\pm$      0.2759&      7.6781 \\\hline
\end{tabular}
\label{table:experiment-parkinsons-motor}
\end{table}

\begin{table}
\caption{Experiment results (test errors) on the Parkinson's -- Total dataset.}
\small
\rowcolors{2}{gray!25}{white}
\begin{tabular}{lrrrrr}
\rowcolor{gray!50}
\hline
$\varphi$ & $n$ & NBR & CSO-NBR(mean) & CSO-NBR(best)\\ 
\hline 
        0.0 &           3 &     18.9802 &      9.6309 $\pm$      0.1369&      9.4488 \\
        0.0 &          10 &     18.9802 &      9.6215 $\pm$      0.2970&   \bf9.1104 \\
        0.0 &          20 &     18.9802 &      9.6947 $\pm$      0.3364&      9.2157 \\
        0.1 &           3 &     18.9802 &   \bf9.5537 $\pm$      0.1006&      9.3631 \\
        0.1 &          10 &     18.9802 &      9.9890 $\pm$      0.1606&      9.7112 \\
        0.1 &          20 &     18.9802 &      9.9324 $\pm$      0.1307&      9.7260 \\
        0.5 &           3 &     18.9802 &      9.5729 $\pm$      0.1130&      9.4543 \\
        0.5 &          10 &     18.9802 &      9.7709 $\pm$      0.2050&      9.4648 \\
        0.5 &          20 &     18.9802 &      9.8207 $\pm$      0.1653&      9.6467 \\
        1.0 &           3 &     18.9802 &      9.5969 $\pm$      0.1199&      9.4083 \\
        1.0 &          10 &     18.9802 &      9.6908 $\pm$      0.2172&      9.2791 \\
        1.0 &          20 &     18.9802 &      9.8576 $\pm$      0.2160&      9.3910 \\\hline
\end{tabular}
\label{table:experiment-parkinsons-total}
\end{table}

\begin{table}
\caption{Experiment results (test errors) on the Appliance Energy dataset.}
\small
\rowcolors{2}{gray!25}{white}
\begin{tabular}{lrrrrr}
\rowcolor{gray!50}
\hline
$\varphi$ & $n$ & NBR & CSO-NBR(mean) & $p$ value & CSO-NBR(best)\\ 
\hline 
        0.0 &           3 &     96.1397 &     95.6540 $\pm$      5.1323&      0.3858&     91.0486 \\
        0.0 &          10 &     96.1397 &     98.6674 $\pm$     10.4532&      0.7680&     90.2488 \\
        0.0 &          20 &     96.1397 &     94.1240 $\pm$      2.0230&      0.0059&     91.7284 \\
        0.1 &           3 &     96.1397 &     92.5080 $\pm$      2.3186&      0.0004&  \bf{88.9964} \\
        0.1 &          10 &     96.1397 &    110.8083 $\pm$     15.0077&      0.9935&     91.6589 \\
        0.1 &          20 &     96.1397 &    119.6089 $\pm$     26.7894&      0.9891&     92.8323 \\
        0.5 &           3 &     96.1397 &     91.2650 $\pm$      0.7496&      0.0000&     89.9534 \\
        0.5 &          10 &     96.1397 &    116.4989 $\pm$     20.3208&      0.9943&     95.5893 \\
        0.5 &          20 &     96.1397 &    118.0088 $\pm$     17.2667&      0.9985&     93.1136 \\
        1.0 &           3 &     96.1397 &\bf{90.9198} $\pm$      1.1368&      0.0000&     89.9421 \\
        1.0 &          10 &     96.1397 &    118.5223 $\pm$      8.2297&      1.0000&    102.9101 \\
        1.0 &          20 &     96.1397 &    123.2761 $\pm$     11.2777&      1.0000&    109.8786 \\
\hline
\end{tabular}
\label{table:experiment-appliance-energy}
\end{table}

\begin{table}
\caption{Experiment results (test errors) on the Survival dataset.}
\small
\rowcolors{2}{gray!25}{white}
\begin{tabular}{lrrrrr}
\rowcolor{gray!50}
\hline
$\varphi$ & $n$ & NBR & CSO-NBR(mean) & $p$ value & CSO-NBR(best)\\ 
\hline 
        0.0 &           3 &      3.2272 &      3.0148 $\pm$      0.2732&      0.0181&      2.6715 \\
        0.0 &          10 &      3.2272 &      2.9216 $\pm$      0.3084&      0.0060&      2.2261 \\
        0.0 &          20 &      3.2272 &   \bf2.7333 $\pm$   \bf0.2110&      0.0000&      2.3555 \\
        0.1 &           3 &      3.2272 &      4.0539 $\pm$      0.9391&      0.9894&      3.5272 \\
        0.1 &          10 &      3.2272 &      3.7990 $\pm$      0.4212&      0.9990&      2.9585 \\
        0.1 &          20 &      3.2272 &      3.5180 $\pm$      0.5446&      0.9372&      2.8164 \\
        0.5 &           3 &      3.2272 &      2.7568 $\pm$      0.4411&      0.0041&   \bf2.1783 \\
        0.5 &          10 &      3.2272 &      3.7682 $\pm$      0.7155&      0.9798&      2.5391 \\
        0.5 &          20 &      3.2272 &      3.1950 $\pm$      0.2570&      0.3505&      2.8708 \\
        1.0 &           3 &      3.2272 &      2.7909 $\pm$      0.3210&      0.0010&      2.3641 \\
        1.0 &          10 &      3.2272 &      3.6973 $\pm$      0.7256&      0.9646&      2.7918 \\
        1.0 &          20 &      3.2272 &      3.4808 $\pm$      0.4700&      0.9389&      2.6984 \\
\hline
\end{tabular}
\label{table:experiment-survival}
\end{table}

Tables \ref{table:experiment-red-wine} to \ref{table:experiment-survival}
summarise the results.
The column names have the same interpretation as previously,
and for each table, the best achieved test set errors 
(both averaged over ten runs and for a single run) 
have been bolded.

Firstly we examined the results
in respect of the parameter values we were varying.
The results show that in most cases,
a low value of $\varphi$ (either 0.0 or 0.1)
achieves the lowest mean test error.
The only exceptions to this are the Red Wine dataset
and the Appliance Energy
dataset for which a higher $\varphi$ of 1.0
leads to the best overall results.
However, runs with lower $\varphi$ still
significantly outperform the baseline
on the Appliance Energy dataset.

The results are not as clear-cut with the dataset size
parameter $n$ however. 
There is no consistent pattern of optimal artificial
dataset size across the problems.
For example, the Survival dataset appears to require
a large artificial dataset with $n=20$.
When $n$ is reduced, performance degrades significantly.
Conversely, for the Energy Efficiency -- Heating
dataset, a very small number of artificial examples (i.e. $n=3$)
is preferred and increasing this reduces accuracy.

The takeaway from the analysis is that 
while there is no clearly optimal
artificial dataset size, a low value of $\varphi$
(either 0.0 or 0.1) 
appears to be a good default in most cases.
Specific problems may benefit from gradually 
increasing $\varphi$ if performance is not ideal.

The next step in our examination
was to perform statistical significance tests
comparing NBR and CSO-NBR for a more rigorous
comparison.
Similar to the previous section, we performed
single sample $t$-tests comparing the test set errors
of CSO-NBR against the single result obtained
by the deterministic NBR algorithm,
for each set of ten runs.
For the first six problems,
the results of the tests unanimously
showed CSO-NBR significantly outperforming NBR
with exceptionally low $p$ values 
(i.e. $p\ll0.0001$),
but higher $p$ values were exhibited 
in the cases of the latter two 
(Appliance Energy and Survival) problems.
We therefore include $p$ values in the tables of results
only for the latter two regression problems,
since these are the interesting results where the significance
varies considerably.

\begin{table}
\caption{Test error of linear regression (LR) 
and parameter-optimised Gaussian processes regression (GPR)
on the datasets used in the experiment.}
\small
\rowcolors{2}{gray!25}{white}
\begin{tabular}{lrr}
\rowcolor{gray!50}
\hline
Dataset & GPR & LR\\ 
\hline
Wine Quality -- Red &0.66&0.68\\
Wine Quality -- White &0.84&0.83\\
Energy Efficiency -- Heating &2.08&3.78\\
Energy Efficiency -- Cooling &2.44&3.54\\
Parkinson's -- Motor &16.20&7.07\\
Parkinson's -- Total &11.97&9.33\\
Appliance Energy &177.20& 92.55\\
Survival &2.47& 5.16\\
\hline 
\end{tabular}
\label{table:lr}
\end{table}

Finally, we also ran standard with default parameters
and parameter optimised GPR on the same datasets.
The test set errors are shown in Table \ref{table:lr},
and are mainly provided for context:
the main focus of our comparison is between NBR and CSO-NBR in
this section.
To summarise,
CSO-NBR significantly improves NBR's performance,
to the point that it is at competitive with,
and sometimes better than, LR,
and occasionally better than state-of-the-art black box methods
such as GPR.

\section{Analysis}\label{section:analysis}

In this section, 
we explore the performance of CSO-NBR 
on a very simple two dimensional dataset with known properties,
in an attempt to understand the interaction between
the original NBR algorithm
and our enhanced version.

\begin{figure}
\includegraphics[width=0.75\textwidth]{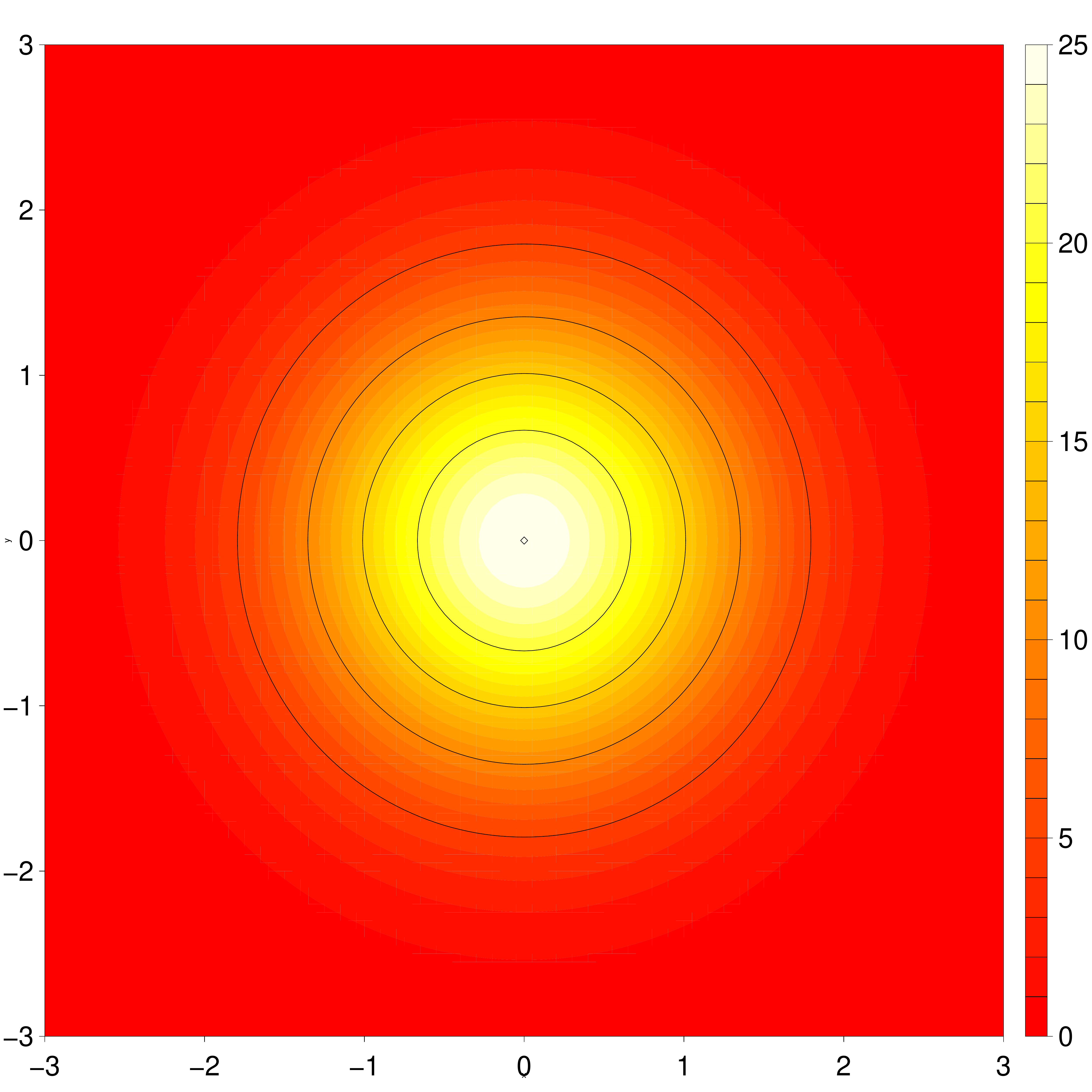}
\caption{Contour plot showing a two-dimensional distribution
from which artificial data points were sampled.
The distribution is defined by the equation $z=25e^{-\frac{1}{2}r^2}$
where $r$ is the distance of each sample point $(x,y)$ from the origin.}
\label{fig:ground_truth}
\end{figure}

The artificial dataset we employ in this section
is generated
using the ``peak'' algorithm in the R package \texttt{MLBench}
\citep{leisch10}.
Figure \ref{fig:ground_truth} depicts the ground truth
from which the data is generated.
The data is drawn from a two dimensional
normal distribution centred on the origin
and scaled to a maximum height of 25.
Points are sampled uniformly from the distribution 
within a circle of radius three about the origin. 
To generate a training dataset, we sampled 100 such points.
A test set was then generated from a further 100 random samples.
Each example in the dataset is a tuple of the form $(x,y,z)$
where $z$ is the target to predict.

We note the following points about this artificial regression problem:
\begin{enumerate}
\item The target variable $z$ is a non-linear function
of the two predictors $x$ and $y$.
This property of the data makes the problem difficult
for simple linear model learners such as LR.
This fact is demonstrated 
by the performance achieved using the LR algorithm
to predict the $z$ values on both the training and test datasets
after model learning using the training data:
LR achieves a very poor RMSE score of 8.4 on the training data
and 8.7 on the test data.
\item The predictive features $x$ and $y$ 
are not conditionally independent given the predictive target $z$,
which violates naive Bayes' conditional independence assumption.
Once $z$ is known, the potential values for $(x,y)$ 
are distributed around the perimeter
of a circle centred on the origin.
Thus, knowing the value of $x$ limits the locations for $y$ to two possible values.
In contrast to LR, NBR achieves RMSE scores of 1.4 and 2.1 
on the two data splits respectively,
which is a major improvement but still not ideal.
\end{enumerate}

\begin{figure}
\includegraphics[width=0.75\textwidth]{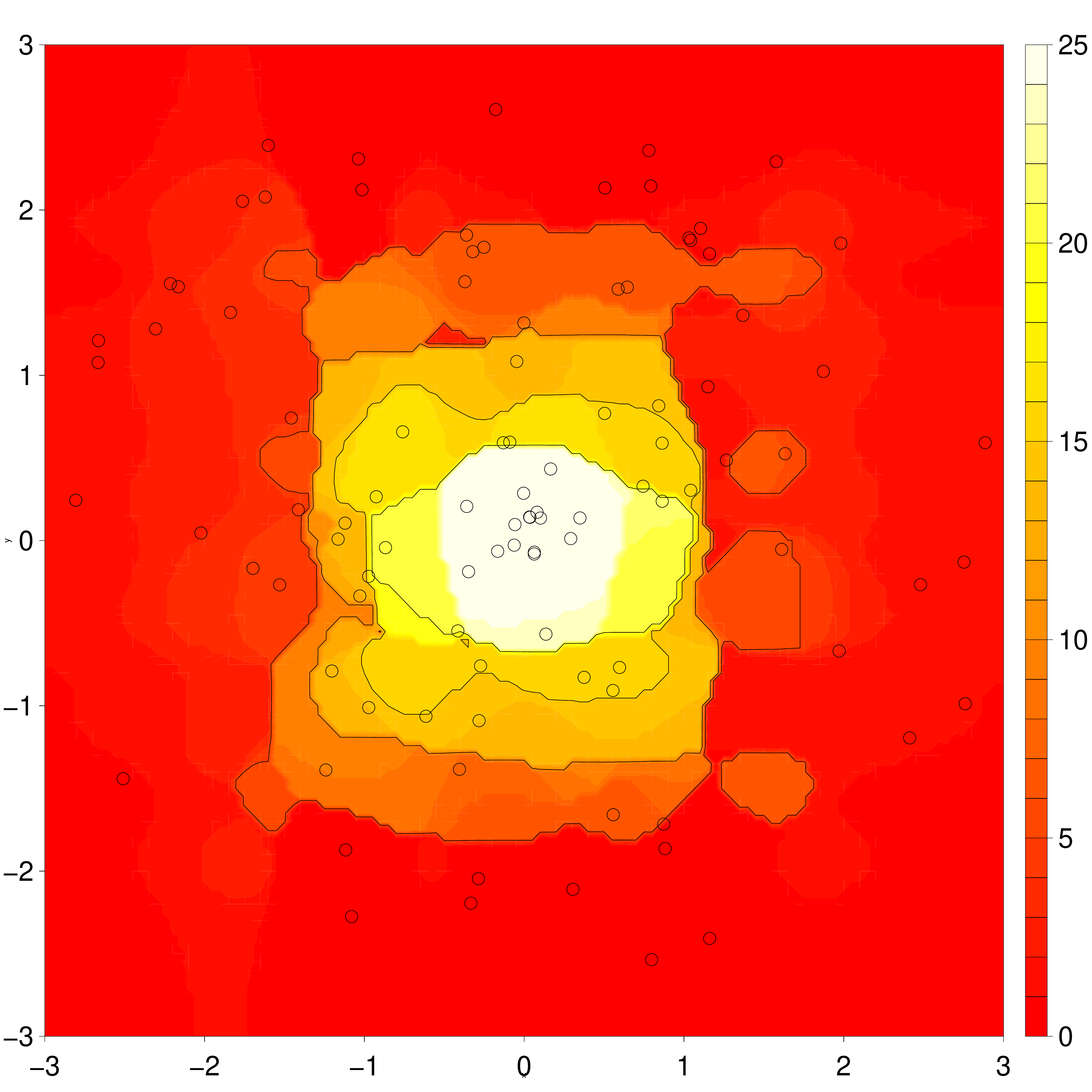}
\caption{Predicted distribution using the NBR algorithm
trained using the 100 training examples.
The examples are depicted in the figure as points,
and were generated using the R package \texttt{MLBench} \citep{leisch10}.}
\label{fig:nbr_predictions}
\end{figure}

To visualise the model that NBR learns from the training data,
we constructed the contour plot depicted in Figure \ref{fig:nbr_predictions}. 
Data from the plot was generated by dividing the $[-3\dots3]\times[-3\dots3]$
square centred on the origin into a $120\times120$ grid.
At each grid point, we used the NBR model to make
a prediction.
Thus, $121\times121$ predictions were made in total,
and the contour plot shows the predicted values.
Figure \ref{fig:nbr_predictions} also depicts 
the 100 randomly sampled training
data points used to construct the model
as points on the contour plot.

Next, we ran the CSO-NBR algorithm ten times
with the same settings as before 
($s=100$, $\varphi=0.1$, $t_{max}=1000$, $n=10$)
and selected the resulting single best artificial training dataset
from across the ten runs.
The average model achieved an RMSE of $0.7\pm0.12$
on the test data.
The best model, on the other hand,
achieved an RMSE score of 0.4 on the training data
and 0.5 on the test data 
-- both the average and best result are a substantial improvement
on both LR and standard NBR.

\begin{figure}
\includegraphics[width=0.75\textwidth]{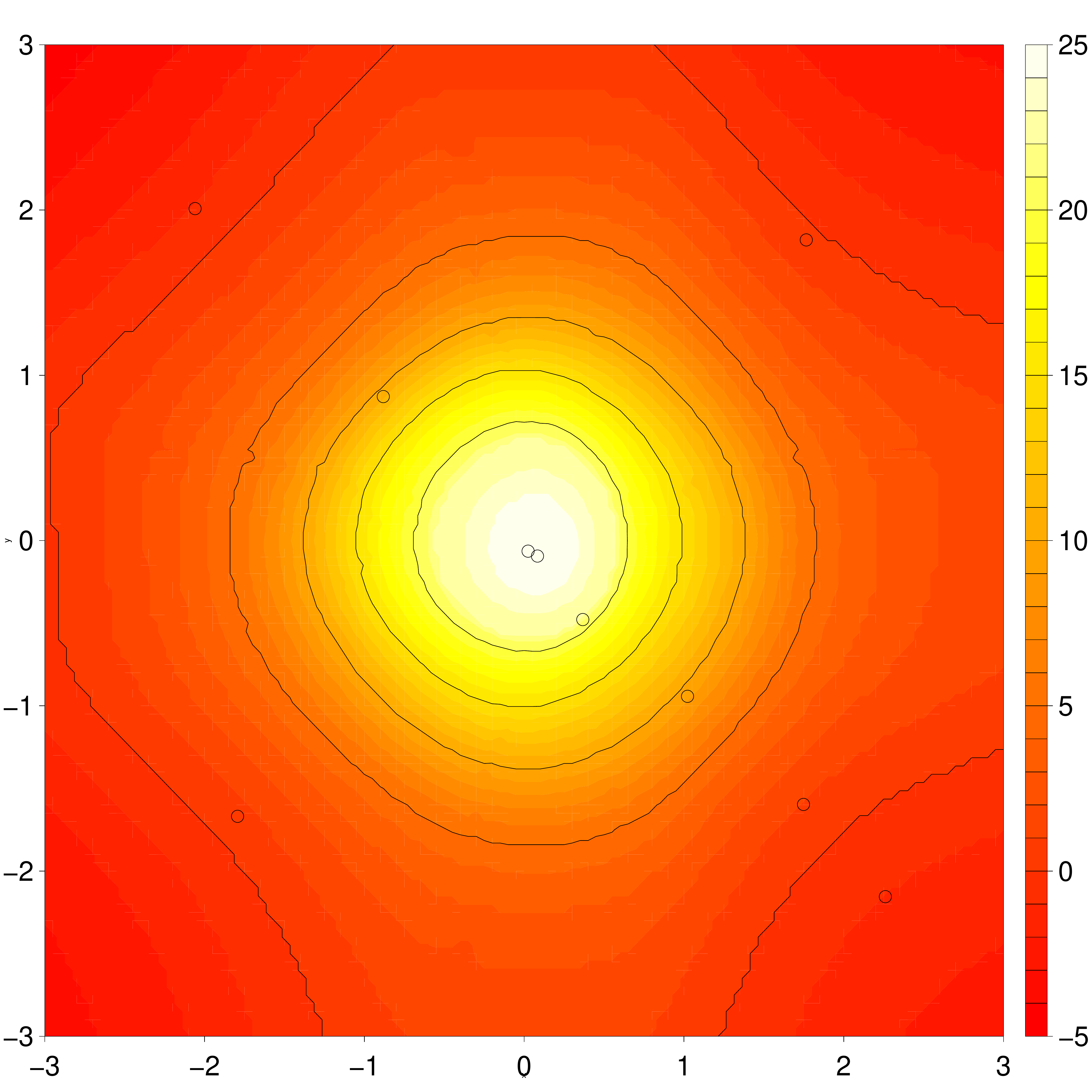}
\caption{Predicted distribution using the CSO-NBR algorithm
trained using ten optimised artificial training examples.
The examples are depicted in the figure as points.}
\label{fig:siad_nbr_predictions}
\end{figure}

\begin{table}
\caption{Artificial data (rounded to 2 d.p.) used to train the model whose predictions
are depicted in Figure \ref{fig:siad_nbr_predictions}.}
\small
\rowcolors{2}{gray!25}{white}
\begin{tabular}{rrr}
\rowcolor{gray!50}
\hline
$x$ & $y$ & $z$\\
\hline
1.75 & -1.60 & 0.72\\
0.08 & -0.09 & 30.89\\
1.02 & -0.94 & 9.87\\
0.02 & -0.06 & -2.81\\
2.26 & -2.15 & -5.22\\
-0.88 & 0.87 & 11.73\\
0.37 & -0.48 & 17.51\\
1.76 & 1.82 & 1.58\\
-2.06 & 2.01 & -2.91\\
\hline
\end{tabular}
\label{table:artificial}
\end{table}

Figure \ref{fig:siad_nbr_predictions} depicts the predictions
made by the CSO-NBR model
along with the $(x,y)$ coordinates of the
ten artificial data points in the best artificial datasets
used to build the model.
Table \ref{table:artificial} further shows the actual
artificial data points themselves.
We can make several observations from the contour plots
of the predictions and the table.
Firstly, as Figure \ref{fig:nbr_predictions} clearly shows,
the NBR algorithm by itself
models the ground truth to a reasonable degree. 
The predicted $z$ values are higher
the closer that $(x,y)$ is to the origin,
which is what is expected.
However, the circular symmetricity of the distribution
in Figure \ref{fig:nbr_predictions} 
is not preserved well, as the contours
have an obvious ``squarish'' appearance.
There also appears to be a high degree of noise
in the contours, suggesting that the model
is not generalising well and instead overfitting
the training data noise.

In contrast, the CSO-NBR predictions
as depicted in Figure \ref{fig:siad_nbr_predictions}
produce a much cleaner, more symmetric, and less noisy
predicted distribution.
The prediction agrees well with the ground
truth distribution in Figure \ref{fig:ground_truth},
which in turn explains the low RMSE scores achieved
in the experiments we performed.
Of interest is the layout of
the artificial data points across the
distribution of predictions.
Clearly, the positions of the artificial
points have been highly optimised 
with the majority of the points lying 
approximately on the negative diagonal
$y=-x$ and the remaining two 
points lying on the opposite diagonal.
This configuration of points,
when used as training data for the NBR algorithm,
produces superior predictions
and indicates that only a small amount of training
data is required to effectively model this problem.

We additionally performed some time experiments
with NBR and CSO-NBR on the artificial data.
Over ten runs on an 
Intel Core i7-6700K CPU running at 4.00GHz with 64GB of RAM
and using the same settings as above, 
the mean training time for NBR was 0.017 seconds,
compared to 98.97 seconds for CSO-NBR.
CSO-NBR made use of six cores while our NBR implementation
is single-threaded.
When we increased the dimensionality of problem from two to ten dimensions
(by creating a new artificial dataset with the same number of examples),
the mean runtimes of both algorithms
was 0.097 seconds vs. 428.23 seconds.
For the 30-dimensional peak problem, the runtimes
were 0.28 seconds and 1241.99 seconds.

Clearly, the training time for CSO-NBR is much higher
than NBR (which is to be expected, since each run of
CSO-NBR builds approximately 500,000 NBR models),
and moreover, the training time increases at a rate that is worse
than linear with the dimensionality of the regression problem.
We expect that the training for CSO-NBR, in practice,
would be even worse,
since in these experiments 
we have held the number of iterations constant.
With an increase in dimensionality, however, comes a corresponding
increase in the size of the search space for CSO,
and therefore to properly solve the problem
CSO will need significantly more iterations.
Mitigating this, however, is the fact that
CSO is an ``embarrassingly parallel" algorithm:
i.e. doubling the number of cores available 
should effectively halve the runtime.

\section{Conclusion}\label{section:conclusion}

In summary,
we have described a method for 
enhancing the predictive accuracy
of naive Bayes for regression.
The approach employs
``real'' training data
only indirectly in the machine learning pipeline,
as part of a fitness function
that in turn is used to optimise a small
artificial surrogate training dataset.
The naive Bayes model 
is constructed using the artificial data
instead of the real training data.
Our evaluation has demonstrated that the method
often produces superior regression models
in terms of generalisation error.

The primary advantages of this idea
are twofold.
In its original formulation, 
naive Bayes typically underperforms other regression models.
However, the algorithm does have certain specific
advantages such as being simple and interpretable, 
and interpretable models are important in many areas
(e.g. medicine)
where understanding the rules used for prediction
is important.
Finding new methods to enhance the accuracy 
of naive Bayesian models is therefore important,
and we have described one means of doing so.

Several interesting questions are raised by this research,
which may form a basis for future work.
For example,
how hard is this problem when framed as an optimisation problem,
from a fitness landscape perspective?
``Easy'' problems in evolutionary computation are usually those
with a high correlation between the fitness of solutions
and their distance to the global optima.
``Hard'' problems, on the other hand, have a much lower or even
negative (in the case of deception) correlation.

Another question concerns the behaviour of our proposed algorithm
when a base algorithm quite different to NBR is employed.
How exactly does CSO's behaviour change with the base model?
Can we replace NBR with a full Bayesian network 
or a decision tree ensemble and still get generalisation improvements?
Do we need to use larger or smaller artificial datasets to make 
our approach work with other base algorithms?

There is also the issue of how to select the optimal
artificial dataset size for the problem at hand.
Our results indicate that there is no single value of $n$
that is a good default in most cases.
Some problems require a larger $n$, and others a smaller $n$.
How, then, can the correct $n$ be chosen?
A potential but expensive answer to this question 
is to run the algorithm multiple times 
with different artificial dataset sizes and attempt
to hone in on the best one.
However, other more intelligent approaches 
(for example, greedily adding one optimised example 
to the artificial dataset at a time
until no more improvements are possible) 
are certainly possible and may work.

Understanding how and why the algorithm works theoretically
is a vital next step.
Recently, efficient algorithms for computing influence
functions have been proposed in the literature  \citep{koh17}.
Influence functions relate the change in weight of a training
example to training loss, and are therefore a means of assessing
the impact of each individual example on predictive performance.
Are these ideas applicable here?
If so, we could use them to understand the relationship
betweens real and fake training data, and model predictions.

Research involving surrogate models for evolutionary algorithms
may also yield improvements.
If the fitness of a model trained from a dataset can be
estimated approximately using a different model, 
then the optimisation speed could potentially speed
up significantly.
Again, influence functions may be useful other;
other surrogate model approaches could also be explored.

Finally, 
is it possible that useful information
might be discovered in these small highly-optimised artificial datasets,
in the same way that rules and other knowledge 
can be typically discovered by inspecting normal machine learning models?
Figure \ref{fig:siad_nbr_predictions} shows that the optimised artificial
data points are uniformly distributed across the space of examples.
Would it be possible to take an artificial dataset like this 
and use a different type of analysis to extract useful knowledge
about the problem domain?

\bibliographystyle{spbasic}      
\bibliography{refs}   

\section*{Appendix}
\begin{table}\caption{Results of running SPSO and SPSO-NBR on the datasets used
in the first experiment with $s=50$ and $t_{max}=1000$.
Shown are the test errors.
Bolded figures in the SPSO-NBR(mean) column indicate a better result
than the corresponding CSO-NBR(mean) result.}
\small
\rowcolors{2}{gray!25}{white}
\begin{tabular}{lrrr}
\rowcolor{gray!50}
\hline
Dataset & NBR & SPSO-NBR(mean) & SPSO-NBR(best)\\ 
\hline 
         auto93  &         5.41  &         7.28  $\pm$         2.53  &         4.46 \\
      autoHorse  &        12.92  &        19.53  $\pm$         2.84  &        16.56 \\
        autoMpg  &         3.43  &         3.56  $\pm$         0.28  &         3.04 \\
      autoPrice  &      2375.45  &      2418.18  $\pm$       217.03  &      2171.16 \\
       baskball  &         0.11  &         0.10  $\pm$         0.01  &         0.08 \\
        bodyfat  &         1.68  &         1.33  $\pm$         1.24  &         0.70 \\
          bolts  &         7.92  &      \bf9.48  $\pm$         2.22  &         5.82 \\
    cholesterol  &        61.76  &     \bf52.80  $\pm$         2.04  &        49.40 \\
      cleveland  &         1.05  &         0.86  $\pm$         0.07  &         0.79 \\
          cloud  &         0.43  &         0.55  $\pm$         0.20  &         0.37 \\
            cpu  &       114.65  &        56.95  $\pm$        32.54  &        19.48 \\
        detroit  &        89.18  &        63.51  $\pm$        21.85  &        38.91 \\
     echoMonths  &        12.67  &     \bf12.20  $\pm$         1.14  &        10.74 \\
        elusage  &        17.73  &        17.84  $\pm$         2.32  &        14.93 \\
      fishcatch  &       209.45  &        80.08  $\pm$        16.97  &        53.79 \\
       fruitfly  &        19.80  &     \bf16.89  $\pm$         0.93  &        15.40 \\
        gascons  &        17.63  &         5.54  $\pm$         2.63  &         3.04 \\
        housing  &         6.32  &         4.53  $\pm$         0.87  &         3.95 \\
      hungarian  &         0.36  &         0.34  $\pm$         0.02  &         0.32 \\
        longley  &       552.15  &       910.21  $\pm$       504.25  &       408.73 \\
         lowbwt  &       559.99  &       479.54  $\pm$        30.36  &       455.66 \\
       mbagrade  &         0.30  &         0.33  $\pm$         0.02  &         0.31 \\
           meta  &       378.52  &       511.75  $\pm$       243.39  &       184.12 \\
            pbc  &      1049.74  &      1026.42  $\pm$        46.36  &       964.19 \\
        pharynx  &       354.80  &       330.98  $\pm$        31.81  &       303.07 \\
      pollution  &        62.31  &        66.89  $\pm$        26.40  &        43.90 \\
       pwLinear  &         2.49  &         2.24  $\pm$         0.17  &         2.04 \\
          quake  &         0.27  &         0.20  $\pm$         0.00  &         0.19 \\
       schlvote  &   1956688.47  &   1222473.32  $\pm$    297248.95  &   1028636.70 \\
        sensory  &         0.94  &         0.82  $\pm$         0.02  &         0.78 \\
          servo  &         1.04  &         0.88  $\pm$         0.18  &         0.67 \\
          sleep  &         4.09  &        11.43  $\pm$        20.98  &         2.43 \\
         strike  &       696.60  &    \bf672.86  $\pm$        10.21  &       656.73 \\
        veteran  &       164.17  &       191.26  $\pm$        45.31  &       145.96 \\
       vineyard  &         1.33  &      \bf2.35  $\pm$         0.25  &         1.91 \\
\hline
\end{tabular}
\label{table:PSO_results1}
\end{table}

\begin{table}\caption{Results of running SPSO and SPSO-NBR on the datasets used
in the first experiment with $s=100$ and $t_{max}=500$.
Shown are the test errors.
Bolded figures in the SPSO-NBR(mean) column indicate a better result
than the corresponding CSO-NBR(mean) result.}
\small
\rowcolors{2}{gray!25}{white}
\begin{tabular}{lrrr}
\rowcolor{gray!50}
\hline
Dataset & NBR & SPSO-NBR(mean) & SPSO-NBR(best)\\ 
\hline 
         auto93  &         5.41  &         6.58  $\pm$         1.23  &         5.07 \\
      autoHorse  &        12.92  &        17.08  $\pm$         2.74  &        12.61 \\
        autoMpg  &         3.43  &         3.55  $\pm$         0.22  &         3.19 \\
      autoPrice  &      2375.45  &      2524.65  $\pm$       536.17  &      2217.93 \\
       baskball  &         0.11  &         0.10  $\pm$         0.04  &         0.08 \\
        bodyfat  &         1.68  &         0.70  $\pm$         0.10  &         0.53 \\
          bolts  &         7.92  &      \bf9.11  $\pm$         3.41  &         5.12 \\
    cholesterol  &        61.76  &     \bf54.37  $\pm$         2.47  &        50.80 \\
      cleveland  &         1.05  &         0.84  $\pm$         0.08  &         0.75 \\
          cloud  &         0.43  &         0.55  $\pm$         0.16  &         0.34 \\
            cpu  &       114.65  &        77.61  $\pm$        36.68  &        25.26 \\
        detroit  &        89.18  &     \bf45.27  $\pm$        11.89  &        28.58 \\
     echoMonths  &        12.67  &     \bf11.89  $\pm$         0.75  &        11.25 \\
        elusage  &        17.73  &     \bf17.35  $\pm$         2.30  &        13.38 \\
      fishcatch  &       209.45  &        92.72  $\pm$        66.22  &        55.98 \\
       fruitfly  &        19.80  &        18.02  $\pm$         1.25  &        16.48 \\
        gascons  &        17.63  &         5.93  $\pm$         2.48  &         2.39 \\
        housing  &         6.32  &         4.10  $\pm$         0.13  &         3.92 \\
      hungarian  &         0.36  &         0.34  $\pm$         0.02  &         0.31 \\
        longley  &       552.15  &      1061.36  $\pm$      1367.85  &       372.86 \\
         lowbwt  &       559.99  &    \bf522.28  $\pm$        59.45  &       466.73 \\
       mbagrade  &         0.30  &         0.34  $\pm$         0.03  &         0.32 \\
           meta  &       378.52  &       566.25  $\pm$       316.02  &        97.00 \\
            pbc  &      1049.74  &   \bf1009.47  $\pm$        57.85  &       919.29 \\
        pharynx  &       354.80  &    \bf325.70  $\pm$        44.74  &       272.44 \\
      pollution  &        62.31  &        57.13  $\pm$        15.40  &        44.13 \\
       pwLinear  &         2.49  &         2.26  $\pm$         0.27  &         2.01 \\
          quake  &         0.27  &         0.20  $\pm$         0.00  &         0.19 \\
       schlvote  &   1956688.47  &   1157391.48  $\pm$    201257.39  &   1023514.27 \\
        sensory  &         0.94  &         0.83  $\pm$         0.02  &         0.80 \\
          servo  &         1.04  &      \bf0.76  $\pm$         0.12  &         0.60 \\
          sleep  &         4.09  &         4.79  $\pm$         3.24  &         2.55 \\
         strike  &       696.60  &    \bf668.92  $\pm$        13.44  &       655.56 \\
        veteran  &       164.17  &    \bf173.83  $\pm$        30.15  &       140.91 \\
       vineyard  &         1.33  &      \bf2.61  $\pm$         0.72  &         2.11 \\
\hline
\end{tabular}
\label{table:PSO_results2}
\end{table}


\end{document}